\documentclass{article}

\usepackage{times}
\usepackage{graphicx} % more modern
\usepackage{subfigure}
\usepackage{stackengine}

\usepackage{natbib}

\usepackage{algorithmic}
\usepackage{algorithm}
\usepackage[algo2e,linesnumbered]{algorithm2e}

\usepackage{url}

\usepackage[accepted]{icml2019}

\usepackage{graphicx}

\usepackage{amsmath}
\usepackage{amsthm}
\usepackage{amssymb}
\usepackage{color}
\usepackage{subfigure}
\usepackage{multirow}
\usepackage{enumitem}

\usepackage{pifont}

\newcommand{\xmark}{\text{\ding{55}}}

\definecolor{mydarkred}{rgb}{0.6,0,0}
\definecolor{mydarkgreen}{rgb}{0,0.6,0}
\usepackage[colorlinks,
linkcolor=mydarkred,
citecolor=mydarkgreen]{hyperref}
\usepackage{url}

\icmltitlerunning{How does Disagreement Help Generalization against Label Corruption?}

\usepackage{array}
\newcolumntype{L}[1]{>{\raggedright\let\newline\\\arraybackslash\hspace{0pt}}m{#1}}
\newcolumntype{C}[1]{>{\centering\let\newline  \\\arraybackslash\hspace{0pt}}m{#1}}
\newcolumntype{R}[1]{>{\raggedleft\let\newline \\\arraybackslash\hspace{0pt}}m{#1}}

\usepackage[mathscr]{euscript}

\begin{document}

\twocolumn[
\icmltitle{How does Disagreement Help Generalization against Label Corruption?}

\icmlsetsymbol{equal}{*}

\begin{icmlauthorlist}
\icmlauthor{Xingrui Yu}{uts}
\icmlauthor{Bo Han}{riken}
\icmlauthor{Jiangchao Yao}{alibaba}
\icmlauthor{Gang Niu}{riken}
\icmlauthor{Ivor W. Tsang}{uts}
\icmlauthor{Masashi Sugiyama}{riken,utokyo}
\end{icmlauthorlist}

\icmlaffiliation{uts}{CAI, University of Technology Sydney}
\icmlaffiliation{riken}{RIKEN-AIP}
\icmlaffiliation{alibaba}{Alibaba Damo Academy}
\icmlaffiliation{utokyo}{University of Tokyo}

\icmlcorrespondingauthor{Xingrui Yu}{xingrui.yu@student.uts.edu.au}
%\icmlcorrespondingauthor{Masashi Sugiyama}{sugi@k.u-tokyo.ac.jp}

\icmlkeywords{Machine Learning, ICML}

\vskip 0.3in
]

\printAffiliationsAndNotice{}

\begin{abstract}
Learning with noisy labels is one of the hottest problems in weakly-supervised learning. Based on memorization effects of deep neural networks, training on small-loss instances becomes very promising for handling noisy labels. This fosters the state-of-the-art approach ``Co-teaching'' that cross-trains two deep neural networks using the small-loss trick. However, with the increase of epochs, two networks converge to a consensus and Co-teaching reduces to the self-training MentorNet. To tackle this issue, we propose a robust learning paradigm called Co-teaching+, which bridges the ``Update by Disagreement'' strategy with the original Co-teaching. First, two networks feed forward and predict all data, but keep prediction disagreement data only. Then, among such disagreement data, each network selects its small-loss data, but back propagates the small-loss data from its peer network and updates its own parameters. Empirical results on benchmark datasets demonstrate that Co-teaching+ is much superior to many state-of-the-art methods in the robustness of trained models.
\end{abstract}

%%%%%%%%%%%%%%%%%%%%%%%%%%%%%%%%%%%%%%%%%%%%%%%%%%%%%%%%%%%%%%%%%%%%%%%%%

\section{Introduction}
\label{sect:introduction}
In weakly-supervised learning, learning with noisy labels is one of the most challenging questions, since noisy labels are ubiquitous in our daily life, such as web queries~\cite{liu2011noise}, crowdsourcing~\cite{welinder2010multidimensional}, medical images~\cite{dganitraining18training}, and financial analysis~\cite{ait2010high}. Essentially, noisy labels are systematically corrupted from ground-truth labels, which inevitably degenerates the accuracy of classifiers. Such degeneration becomes even more prominent for deep learning models (e.g., convolutional and recurrent neural networks), since these complex models can fully memorize noisy labels \cite{zhang2016understanding,arpit2017closer}.

To handle noisy labels, classical approaches focus on either adding regularization \cite{miyato2016virtual} or estimating the label transition matrix \cite{patrini2017making}. Specifically, both explicit and implicit regularizations leverage the regularization bias to overcome the label noise issue. Nevertheless, they introduced a permanent regularization bias, and the learned classier barely reaches the optimal performance. Meanwhile, estimating the label transition matrix does not introduce the regularization bias, and the accuracy of classifiers can be improved by such accurate estimation. However, the label transition matrix is hard to be estimated, when the number of classes is large.

Recently, a promising way of handling noisy labels is to train on small-loss instances \cite{jiang2018mentornet,ren2018learning}. These works try to select small-loss instances, and then use them to update the network robustly. Among those works, the representative methods are MentorNet \cite{jiang2018mentornet} and Co-teaching \cite{han2018coteaching}. For example, MentorNet pre-trains an extra network, and then it uses the extra network for selecting clean instances to guide the training of the main network. When the clean validation data is not available, self-paced MentorNet has to use a predefined curriculum (e.g., small-loss instances). Nevertheless, the idea of self-paced MentorNet is similar to the self-training approach, and it inherits the same inferiority of accumulated error.

To solve the accumulated error issue in MentorNet, Co-teaching has been developed, which simultaneously trains two networks in a symmetric way \cite{han2018coteaching}. First, in each mini-batch data, each network filters noisy (i.e., big-loss) samples based on the memorization effects. Then, it teaches the remaining small-loss samples to its peer network for updating the parameters, since the error from noisy labels can be reduced by peer networks mutually. From the initial training epoch, two networks having different learning abilities can filter different types of error. However, with the increase of training epochs, two networks will converge to a consensus gradually and Co-teaching reduces to the self-training MentorNet in function.

To address the consensus issue in Co-teaching, we should consider how to always keep two networks diverged within the training epochs, or how to slow down the speed that two networks will reach a consensus with the increase of epochs. Fortunately, we find that a simple strategy called ``Update by Disagreement'' \cite{malach2017decoupling} may help us to achieve the above target. This strategy conducts updates only on selected data, where there is a prediction disagreement between two classifiers.

\begin{figure}[!tp]
\centering
\includegraphics[width=0.45\textwidth]{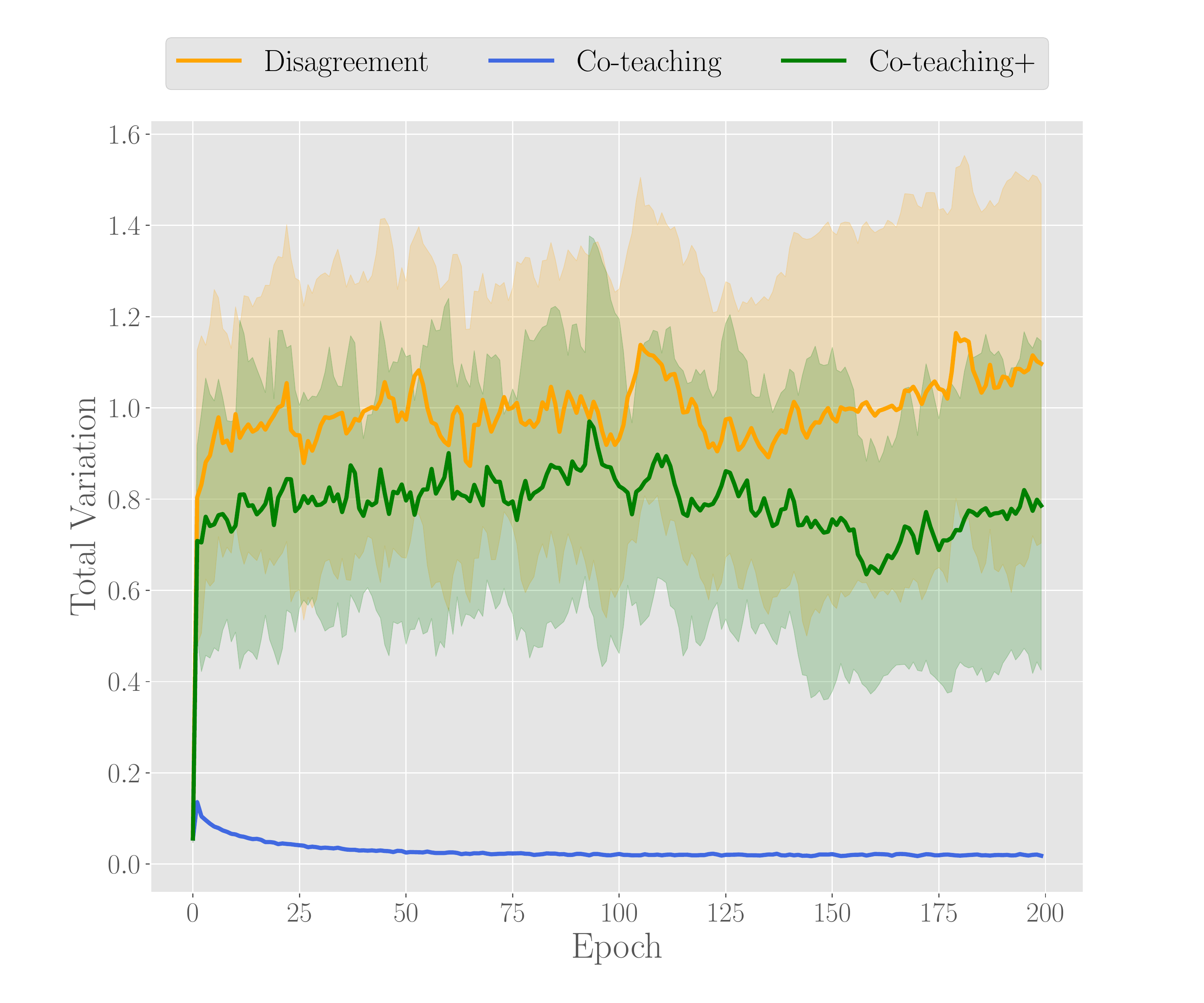}
\caption{Comparison of divergence (evaluated by Total Variation) between two networks trained by the ``Disagreement'' strategy, Co-teaching and Co-teaching+, respectively. Co-teaching+ naturally bridges the ``Disagreement'' strategy with Co-teaching.}
\label{fig:intuition-divergence}
\end{figure}

To demonstrate that the ``Disagreement'' strategy can keep two networks diverged during training, we train two 3-layer MLPs \cite{goodfellow2016deep} on \textit{MNIST} simultaneously for 10 trials, and report total variations of Softmax outputs between two networks in Figure \ref{fig:intuition-divergence}. We can clearly observe that two networks trained by Co-teaching (blue in Figure~\ref{fig:intuition-divergence}) converge to a consensus gradually, while two networks trained by the ``Disagreement'' strategy (orange in Figure~\ref{fig:intuition-divergence}) often keep diverged.

\begin{figure}[!tp]
\centering
%\vspace{-15px}
\includegraphics[width=0.45\textwidth]{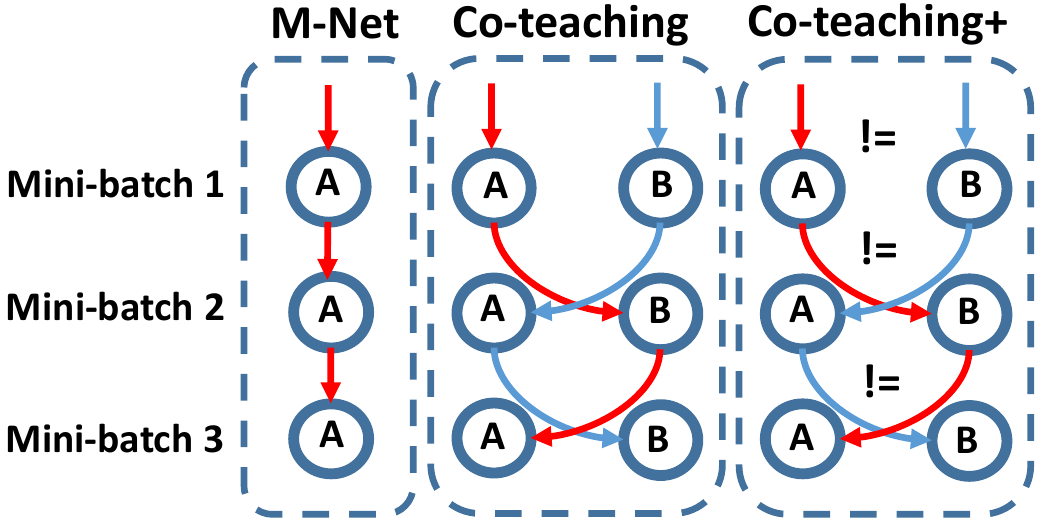}
\caption{Comparison of error flow among MentorNet (M-Net), Co-teaching and Co-teaching+. Assume that the error flow comes from the selection of training instances, and the error flow from network A or B is denoted by red arrows or blue arrows, respectively. \textbf{Left panel:} M-Net maintains only one network (A). \textbf{Middle panel:}
Co-teaching maintains two networks (A \& B) simultaneously. In each mini-batch data, each network selects its small-loss data to teach its peer network for the further training. \textbf{Right panel:} Co-teaching+ also maintains two networks (A \& B). However, two networks feed forward and predict each mini-batch data first, and keep prediction disagreement data (!=) only. Based on such disagreement data, each network selects its small-loss data to teach its peer network for the further training.}
%\vspace{-15px}
\label{fig:intuition-comp}
\end{figure}

Motivated by this phenomenon, in this paper, we propose a robust learning paradigm called Co-teaching+ (Figure~\ref{fig:intuition-comp}), which naturally bridges the ``Disagreement'' strategy with Co-teaching. Co-teaching+ trains two deep neural networks similarly to the original Co-teaching, but it consists of the disagreement-update step (data update) and the cross-update step (parameters update). Initially, in the disagreement-update step, two networks feed forward and predict all data first, and only keep prediction disagreement data. This step indeed keeps two networks (trained by Co-teaching+) diverged (green in Figure~\ref{fig:intuition-divergence}). Then, in the cross-update step, each network selects its small-loss data from such disagreement data, but back propagates the small-loss data from its peer network and updates its own parameters. Intuitively, the idea of disagreement-update comes from Co-training \cite{blum1998combining}, where two classifiers should keep diverged to achieve the better ensemble effects. The intuition of cross-update comes from culture evolving hypothesis \cite{bengio2014evolving}, where a human brain can learn better if guided by the signals produced by other humans.

We conduct experiments on both simulated and real-world noisy datasets, including noisy \textit{MNIST}, \textit{CIFAR-10}, \textit{CIFAR-100}, \textit{NEWS}, \textit{T-ImageNet} and three \textit{Open-sets} \cite{wang2018iterative}. Empirical results demonstrate that the robustness of deep models trained by the Co-teaching+ approach is superior to many state-of-the-art methods, including Co-teaching, MentorNet and F-correction \cite{patrini2017making}. Before delving into details, we clearly emphasize our contribution as follows.
\vspace{-5px}
\begin{itemize}
  \itemsep0em
  \item We denote that ``Update by Disagreement'' (i.e., the Decoupling algorithm) itself \textit{cannot} handle noisy labels, which has been empirically justified in Section~\ref{sect:experiments}.
  \item We realize that the ``Disagreement'' strategy \textit{can} keep two networks diverged, which significantly boosts the performance of Co-teaching.
  \item We summarize \textit{three} key factors towards training robust deep networks with noisy labels: (1) using the small-loss trick; (2) cross-updating parameters of two networks; and (3) keeping two networks diverged.
\end{itemize}

The rest of this paper is organized as follows. In Section~\ref{sect:coteaching+}, we propose our robust learning paradigm Co-teaching+. Experimental results are discussed in Sections~\ref{sect:experiments} and \ref{sect:experiments_openset}. Conclusions are given in Section~\ref{sect:conclusion}.

\section{Co-teaching+: Towards Training of Robust Deep Networks with Noisy Labels}
\label{sect:coteaching+}

Similar to Co-teaching, we also train two deep neural networks. As in Figure~\ref{fig:intuition-comp}, in each mini-batch data, each network conducts its own prediction, then selects instances for which there is a prediction disagreement between two networks. Based on such disagreement data, each network further selects its small-loss data, but back propagates the small-loss data selected by its peer network and updates itself parameters. We call such algorithm as Co-teaching+ (Algorithm~\ref{alg:coteaching+}), which consists of disagreement-update step and cross-update step. This brings the question as follows.

\paragraph{How does disagreement benefit Co-teaching?}
To answer this question, we should first understand the main drawback of Co-teaching. In the early stage of training, the divergence of two networks mainly comes from different (random) parameter initialization. Intuitively, this divergence between two networks pushes Co-teaching to become more robust than self-paced MentorNet, since two diverged networks have different abilities to filter different types of error. However, with the increase of training epochs, two networks will gradually converge to be close to each other (blue in Figure~\ref{fig:intuition-divergence}). Thus, Co-teaching degenerates to self-paced MentorNet, and will not promote the learning ability to select clean data any more. To overcome this issue, we need to keep the constant divergence between two networks or slow down the speed that two networks reach a consensus. This intuition comes from Co-training algorithm, where in semi-supervised learning \cite{chapelle2009semi}, the better ensemble effects require to keep diverged more between two classifiers.

Fortunately, the ``Disagreement'' strategy \cite{malach2017decoupling} can help us to keep two networks diverged (orange in Figure~\ref{fig:intuition-divergence}), since this strategy conducts algorithm updates only on selected data, where there is a prediction disagreement between the two classifiers. Therefore, within the whole training epochs, if two networks always select the disagreement data for further training, the divergence of two networks will be always maintained. Specifically, during the training procedure of Co-teaching, if we use the ``Disagreement'' strategy to keep two networks diverged, then we can prevent Co-teaching reducing to self-training MentorNet in function. This brings us the new robust training paradigm Co-teaching+ (Algorithm~\ref{alg:coteaching+}, green in Figure~\ref{fig:intuition-divergence}).

Take ``complementary peer learning'' as an illustrative example for Co-teaching+. When students prepare for their exams, the peer learning will normally more boost their review efficiency than the solo learning. However, if two students are identically good at math but not good at literature, their review process in literature will have no any progress. Thus, the optimal peer should be complementary, which means that a student who is good at math should best review with another student who is good at literature. This point also explains why the diverged peer has more powerful learning ability than the identical peer.

\begin{algorithm}[!tp]\small
1: {\bfseries Input} $w^{(1)}$ and $w^{(2)}$, training set $\mathcal{D}$, batch size $B$, learning rate $\eta$, estimated noise rate $\tau$, epoch $E_{k}$ and $E_{\max}$;

\For{$e = 1,2,\dots,E_{\max}$}{
	
	2: {\bfseries Shuffle} $\mathcal{D}$ into $\frac{|\mathcal{D}|}{B}$ mini-batches; \hfill //noisy dataset
	
	\For{$n = 1,\dots,\frac{|\mathcal{D}|}{B}$}
	{	
		3: {\bfseries Fetch} $n$-th mini-batch $\mathcal{\bar{D}}$ from $\mathcal{D}$;
		
		4: {\bfseries Select} prediction disagreement $\mathcal{\bar{D}}'$ by Eq. \eqref{eq:disagreement};
		
		5: {\bfseries Get} \mbox{$\mathcal{\bar{D}}^{'(1)} = \arg\min_{\mathcal{D}':|\mathcal{D}'|\ge \lambda(e)|\mathcal{\bar{D}}'|}\ell(\mathcal{D}';w^{(1)})$}; \hfill //sample $\lambda(e)\%$ small-loss instances
		
		6: {\bfseries Get} \mbox{$\mathcal{\bar{D}}^{'(2)} = \arg\min_{\mathcal{D}':|\mathcal{D}'|\ge \lambda(e)|\mathcal{\bar{D}}'|}\ell(\mathcal{D}';w^{(2)})$}; \hfill //sample $\lambda(e)\%$ small-loss instances

		7: {\bfseries Update} \mbox{$w^{(1)} = w^{(1)} - \eta\nabla \ell(\mathcal{\bar{D}}^{'(2)};w^{(1)})$}; \hfill //update $w^{(1)}$ by $\mathcal{\bar{D}}^{'(2)}$;
		
		8: {\bfseries Update} \mbox{$w^{(2)} = w^{(2)} - \eta\nabla \ell(\mathcal{\bar{D}}^{'(1)};w^{(2)})$}; \hfill //update $w^{(2)}$ by $\mathcal{\bar{D}}^{'(1)}$;
	}

	9: {\bfseries Update} $\lambda(e)=1-\min\{\frac{e}{E_{k}}\tau, \tau\}$ or $1-\min\{\frac{e}{E_{k}}\tau, (1+\frac{e-E_k}{E_{\max}-E_{k}})\tau\}$;
	
}

10: {\bfseries Output $w^{(1)}$ and $w^{(2)}$.}
\caption{Co-teaching+. Step 4: disagreement-update; Step 5-8: cross-update.}
\label{alg:coteaching+}
\end{algorithm}

\paragraph{Algorithm description.} Algorithm~\ref{alg:coteaching+} consists of the disagreement-update step (step 4) and the cross-update step (step 5-8), where we train two deep neural networks in a mini-batch manner.

In step 4, two networks feed forward and predict the same mini-bach of data $\mathcal{\bar{D}}$=$\{(x_1, y_1), (x_2, y_2), \cdots, (x_B, y_B)\}$ first, where the batch size is $B$. Then, they keep prediction disagreement data $\mathcal{\bar{D}}'$ (Eq. \eqref{eq:disagreement}) according to their predictions $\{\bar{y}_1^{(1)}, \bar{y}_2^{(1)}, \ldots, \bar{y}_B^{(1)} \}$ (predicted by $w^{(1)}$) and $\{\bar{y}_1^{(2)}, \bar{y}_2^{(2)}, \ldots, \bar{y}_B^{(2)} \}$ (predicted by $w^{(2)}$):
\begin{equation}
\label{eq:disagreement}
    \mathcal{\bar{D}}'=\{(x_i, y_i): \bar{y}_i^{(1)} \neq \bar{y}_i^{(2)}\},
\end{equation}
where $i \in \{1,\ldots,B\}$. The intuition of this step comes from Co-training, where two classifiers should keep diverged to achieve the better ensemble effects.

In step 5-8, from the disagreement data $\mathcal{\bar{D}}'$, each network $w^{(1)}$ (resp. $w^{(2)}$) selects its own small-loss data $\mathcal{\bar{D}}^{'(1)}$ (resp. $\mathcal{\bar{D}}^{'(2)}$), but back propagates the small-loss data $\mathcal{\bar{D}}^{'(1)}$ (resp. $\mathcal{\bar{D}}^{'(2)}$) to its peer network $w^{(2)}$ (resp. $w^{(1)}$) and updates parameters. The intuition of step 5-8 comes from the aforementioned culture evolving hypothesis \cite{bengio2014evolving}, where a human brain can learn better if guided by the signals produced by other humans.

In step 9, we update $\lambda(e)$, which controls how many small-loss data should be selected in each training epoch. Due to the memorization effects, deep networks will fit clean data first and then gradually over-fit noisy data.

Thus, at the beginning of training, we keep more small-loss data (with a large $\lambda(e)$) in each mini-batch, which is equivalent to dropping less data. Since deep networks will fit clean data first, noisy data do not matter at the initial training epochs. With the increase of epochs, we keep less small-loss data (with a small $\lambda(e)$) in each mini-batch. As deep networks will over-fit noisy data gradually, we should drop more data. The gradual decrease of $\lambda(e)$ prevents deep networks over-fitting noisy data to some degree.

Similar to Co-teaching, we decrease $\lambda(e)$ quickly at the first $E_{k}$ epochs to stop networks over-fitting to the noisy data, namely $\lambda(e)=1-\frac{e}{E_{k}}\tau$. However, after $E_{k}$ epochs, Co-teaching+ has two types of $\lambda(e)$. The first type keeps a constant $\lambda(e)$, where $\lambda(e) = 1-\tau$; while the second type further decreases $\lambda(e)$ slowly, where $\lambda(e)=1- (1+\frac{e-E_k}{E_{\max}-E_{k}})\tau$. We take an example to explain the difference.

Assume that the estimated noise rate $\tau$ is $30\%$. It means that, after $E_{k}$ epochs, the first type will constantly fetch $70\%$ small-loss data in each mini-batch as ``clean'' data. However, the $\tau$ estimation tends to be inaccurate in practice. Therefore, given the estimated $\tau$, we should fetch less data, e.g., 60\% small-loss data, to keep remained data more clean. This explains why, in real-world noisy datasets, Co-teaching+ chooses the second type to further decrease $\lambda(e)$ slowly after $E_{k}$ epochs (Section~\ref{sect:experiments_openset}).

\begin{table*}[!tp]
	\centering
	\caption{Comparison of state-of-the-art and related techniques with our Co-teaching+ approach.
		In the first column,
		``small loss'': regarding small-loss samples as ``clean'' samples, which is based on the memorization effects of deep neural networks;
		``double classifiers'': training two classifiers simultaneously;
		``cross update'': updating parameters in a cross manner instead of a parallel manner;
        ``divergence'': keeping two classifiers diverged during the whole training epochs.
}
	\label{tab:comp}
	\scalebox{1}
	{
		\begin{tabular}{c | c | c | c | c | c }
			\hline
			& MentorNet & Co-training & Co-teaching & Decoupling & Co-teaching+ \\ \hline
			small loss   & $\checkmark$                          & $\xmark$                              & $\checkmark$                  & $\xmark$                                      & $\checkmark$                        \\ \hline
			double classifiers   & $\xmark$                          & $\checkmark$                              & $\checkmark$                          & $\checkmark$                           & $\checkmark$                              \\ \hline
			cross update &   $\xmark$                    & $\checkmark$                              & $\checkmark$            & $\xmark$                               & $\checkmark$                             \\ \hline

           divergence & $\xmark$                     & $\checkmark$                           & $\xmark$               & $\checkmark$                           & $\checkmark$                             \\ \hline
		\end{tabular}
	}
%\vspace{-10px}
\end{table*}

\paragraph{Relations to other approaches.}
We compare our Co-teaching+ with related approaches in Table~\ref{tab:comp}. We try to find the connections among them, and pinpoint the key factors that can handle noisy labels. First, self-paced MentorNet \cite{jiang2018mentornet} employs the small-loss trick to handle noisy labels. However, this idea is similar to the self-training approach, and it inherits the same inferiority of accumulated error caused by the sample-selection bias. Inspired by Co-training \cite{blum1998combining} that trains double classifiers and cross updates parameters, Co-teaching \cite{han2018coteaching} has been developed to cross train two deep networks, which addresses the accumulated error issue in MentorNet. Note that, Co-training does not exploit the memorization in deep neural networks, while Co-teaching does (i.e., leveraging small-loss trick).

However, with the increase of training epochs, two networks trained by Co-teaching will converge to a consensus, and Co-teaching will reduce to the self-training MentorNet. This brings us to think how to address the consensus issue in Co-teaching. Although Decoupling algorithm \cite{malach2017decoupling} (i.e., ``Update by Disagreement'') itself \textit{cannot} combat with noisy labels effectively, which has been empirically justified in Section~\ref{sect:experiments}, we clearly realize that the ``Disagreement'' strategy can always keep two networks diverged. Such divergence effects can boost the performance of Co-teaching and bring us Co-teaching+, since the better ensemble effects require to keep diverged more between two classifiers due to Co-training.

To sum up, there are three key factors that can contribute to effectively handle noisy labels (first column of Table~\ref{tab:comp}). First, we should leverage the memorization effects of deep networks (i.e., the small-loss trick). Second, we should train two deep networks simultaneously, and cross update their parameters. Last but not least, we should keep two deep networks diverged during the whole training epochs.

\section{Experiments on Simulated Noisy Datasets}
\label{sect:experiments}
%In Section~\ref{sec:setup}, we introduce the experimental setup from six aspects. Section~\ref{sec:comp} provides the empirical results on three noisy benchmark datasets.

%\footnote{+ Figure \ref{fig:20news}(a): We should explain why label precision drops while the accuracy can keep.}
\subsection{Experimental setup}\label{sec:setup}

\paragraph{Datasets.} First, we verify the efficacy of our approach on four benchmark datasets (Table~\ref{tab:dataset}), including three vision datasets (i.e., \textit{MNIST}, \textit{CIFAR-10}, and \textit{CIFAR-100}) and one text dataset (i.e., \textit{NEWS}). Then, we verify our approach on a larger and harder dataset called \textit{Tiny-ImageNet} (abbreviated as \textit{T-ImageNet}) \footnote{https://tiny-imagenet.herokuapp.com/}. These datasets are popularly used for the evaluation of learning with noisy labels in the literature~\cite{reed2014training, goldberger2016training, kiryo2017positive}.

\begin{table}[ht]\small
\centering
\vspace{-10px}
\caption{Summary of data sets used in the experiments.}
\footnotesize
\begin{tabular}{c | c | c | c | c}
	\hline
	         & \# of train & \# of test & \# of class & size  \\ \hline
	 \textit{MNIST}   & 60,000         & 10,000   & 10    & 28$\times$28 \\ \hline
	\textit{CIFAR-10}  & 50,000         & 10,000   & 10     & 32$\times$32 \\ \hline
	\textit{CIFAR-100}  & 50,000         & 10,000   & 100     & 32$\times$32 \\ \hline
	\textit{NEWS} & 11,314         & 7,532   & 7    & 1000-D \\ \hline
	\textit{T-ImageNet} & 100, 000       & 10, 000  & 200    & 64$\times$64 \\ \hline
\end{tabular}
\label{tab:dataset}
\end{table}

Since all datasets are clean, following~\cite{reed2014training,patrini2017making},
we need to corrupt these datasets manually by the label transition matrix $Q$, where $Q_{ij} = \Pr(\tilde{y} = j| y = i)$ given that noisy $\tilde{y}$ is flipped from clean $y$. Assume that the matrix $Q$ has two representative structures:
%\footnote{\checkmark +++ are there any references adopt these two noise types? Q: pairflip is my thinking, seems no any references for use, but an obvious application is the fine-grained classification, and you may make mistake within very similar case.}
(1) Symmetry flipping~\cite{van2015learning}; (2) Pair flipping~\cite{han2018coteaching}: a simulation of fine-grained classification with noisy labels, where labelers may make mistakes only within very similar classes.

\vspace{-5px}
\paragraph{Baselines.}
We compare Co-teaching+ (Algorithm~\ref{alg:coteaching+}) with the following state-of-art approaches, and implement all methods with default parameters by PyTorch, and conduct all the experiments on a NVIDIA Titan Xp GPU.
%and they belong to three orthogonal directions (i.e., (A)-(C)).
%(A). Training on selected samples:

(i). MentorNet~\cite{jiang2018mentornet}.
An extra teacher network is pre-trained and then used to filter out noisy instances for its student network to learn robustly under noisy labels.
Then,
student network is used for classification. We used self-paced MentorNet in this paper;

(ii). Co-teaching~\cite{han2018coteaching}, which trains two networks simultaneously and cross-updates parameters of peer networks. This method can deal with a large number of classes and is more robust to extremely noisy labels;

(iii). Decoupling~\cite{malach2017decoupling}, which updates the parameters only using the instances which have different prediction from two
classifiers.

(iv). F-correction~\cite{patrini2017making}, which corrects the prediction by the label transition matrix. As suggested by the authors, we first train a standard network to estimate the transition matrix $Q$.

(v). As a simple baseline, we compare Co-teaching+ with the standard deep network that directly trains on noisy datasets (abbreviated as Standard).

\paragraph{Network structure.} For \textit{MNIST}, we use a 2-layer MLP. For \textit{CIFAR-10}, we use a network architecture with 2 convolutional layers and 3 fully connected layers. For \textit{CIFAR-100}, the 7-layer network architecture in our paper follows \cite{wang2018iterative}. For \textit{NEWS}, we borrowed the pre-trained word embeddings from GloVe~\cite{pennington2014glove}, and a 3-layer MLP is used with Softsign active function. For \textit{T-ImageNet}, we use a 18-layer Pre-act ResNet \cite{he2016identity}. The network structure here is standard test bed for weakly-supervised learning, and the details are in Table~\ref{tab:netstuc}.

%(B). Estimating the noise transition matrix:
%(C). Regularization: Virtual Adversarial Training (VAT) \cite{miyato2016virtual,miyato2018virtual}, which is based on virtual adversarial loss: a new measure of local smoothness of the
%conditional label distribution given input. Virtual adversarial loss is defined as the robustness of the conditional label distribution around each input data point against local perturbation, which can be used for semi-supervised learning and learning with noisy labels.

\paragraph{Optimizer.} Adam optimizer (momentum=0.9) is with an initial learning rate of 0.001, and the batch size is set to 128 and we run 200 epochs. The learning rate is linearly decayed to zero from 80 to 200 epochs. As deep networks are highly nonconvex, even with the same network and optimization method, different initializations can lead to different local optimal. Thus, following \cite{malach2017decoupling}, we also take two networks with the same architecture but different initializations as two classifiers.

\begin{table*}[!h]\small
	\centering
	\caption{MLP and CNN models used in our experiments on \textit{MNIST}, \textit{CIFAR-10}, \textit{CIFAR-100}/\textit{Open-sets}, and \textit{NEWS}.}
	\label{tab:netstuc}
	\scalebox{0.84}
	{
		\begin{tabular}{c | c | c | c}
			\hline
			MLP on \textit{MNIST} & CNN on \textit{CIFAR-10} & CNN on \textit{CIFAR-100}/\textit{Open-sets} & MLP on \textit{NEWS} \\ \hline
			28$\times$28 Gray Image & 32$\times$32 RGB Image & 32$\times$32 RGB Image & 1000-D Text  \\ \hline
			& & 3$\times$3 Conv, 64 BN, ReLU  & 300-D Embedding  \\
			& 5$\times$5 Conv, 6 ReLU  & 3$\times$3 Conv, 64 BN, ReLU  & Flatten $\rightarrow$ 1000$\times$300 \\
			& 2$\times$2 Max-pool & 2$\times$2 Max-pool & Adaptive avg-pool $\rightarrow$ 16$\times$300 \\
            \cline{2-4}

		    & & 3$\times$3 Conv, 128 BN, ReLU  & \\
		Dense 28$\times$28 $\rightarrow$ 256, ReLU	& 5$\times$5 Conv, 16 ReLU & 3$\times$3 Conv, 128 BN, ReLU  &   Dense 16$\times$300 $\rightarrow$ 4$\times$300  \\
			& 2$\times$2 Max-pool & 2$\times$2 Max-pool &   BN, Softsign \\
            \cline{2-4}

			&  & 3$\times$3 Conv, 196 BN, ReLU  &   \\
			& Dense 16$\times$5$\times$5 $\rightarrow$ 120, ReLU & 3$\times$3 Conv, 196 BN, ReLU  &   Dense 4$\times$300 $\rightarrow$ 300 \\
			& Dense 120 $\rightarrow$ 84, ReLU & 2$\times$2 Max-pool &    BN, Softsign  \\
			\hline

        Dense 256 $\rightarrow$ 10 & Dense 84 $\rightarrow$ 10 & Dense 256 $\rightarrow$ 100/10 &  Dense 300 $\rightarrow$ 7\\ \hline
		\end{tabular}
	}
\end{table*}

\paragraph{Initialization.}
Assume that the noise rate $\tau$ is known. To conduct a fair comparison in benchmark datasets, we set the ratio of small-loss samples $\lambda(e)$ as identical as Co-teaching:
\begin{equation}
    \lambda(e)=1-\min\{\frac{e}{E_{k}}\tau, \tau\},
\end{equation}
where $E_k = 10$.

If $\tau$  is not known in advanced, $\tau$ can be inferred using validation sets~\cite{liu2016classification,yu2018efficient}.
%The choices of $R(T)$ and $\tau$ are analyzed in Section~\ref{sec:rtau}.
%\footnote{? +++ How the decay rate change to influent the final results? In addition, can we over drop some instances (the trick) to overcome the overfit problems? Namely, first drop over noise rate, and then keep a bit, fall back the noise rate, keep constantly. A: I am not sure exactly, but our intuition seems promising to combat with overfitting.}
Note that $\lambda(e)$ only depends on the memorization effect of deep networks but not any specific datasets.

%and we will show such rule works well in all experiments later.
\paragraph{Measurement.}
To measure the performance,
we use the test accuracy,
i.e.,
\textit{test accuracy = (\# of correct predictions) / (\# of test dataset)}. Intuitively, higher test accuracy means that the algorithm is more robust to the label noise.
%Besides,
%we also use the label precision in each mini-batch,
%i.e.,
%\textit{label Precision = (\# of clean labels) / (\# of all selected labels)}.
% to justify why the test accuracy of our Co-teaching is higher than other baselines.
%Specifically, we sample $\lambda(e)$ of small-loss instances in each mini-batch, and then calculate the ratio of clean labels in the small-loss instances.
%Intuitively, higher label precision means less noisy instances in the mini-batch after sample selection, and the algorithm with higher label precision is also more robust to the label noise.

\subsection{Comparison with the State-of-the-Arts}\label{sec:comp}

\begin{figure*}[ht]
\centering\stackunder{\includegraphics[width=0.9\textwidth]{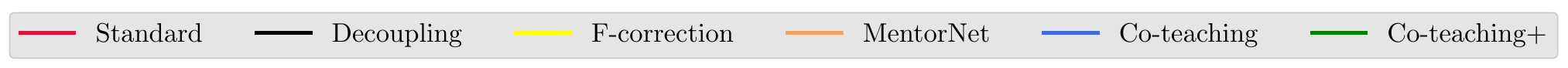}}{}
\subfigure[Pair-45\%.]
{\includegraphics[width=0.32\textwidth]{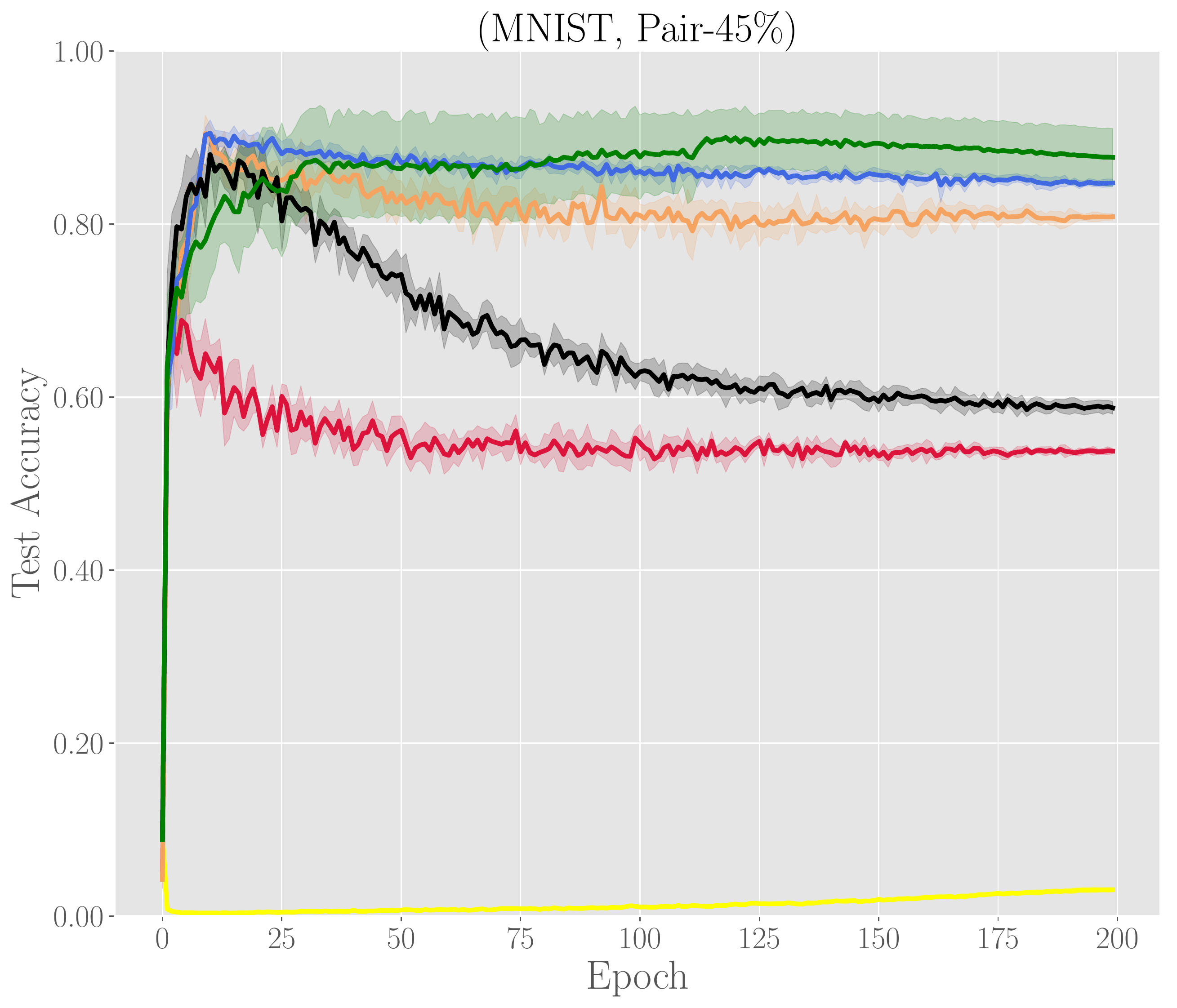}}
\subfigure[Symmetry-50\%.]
{\includegraphics[width=0.32\textwidth]{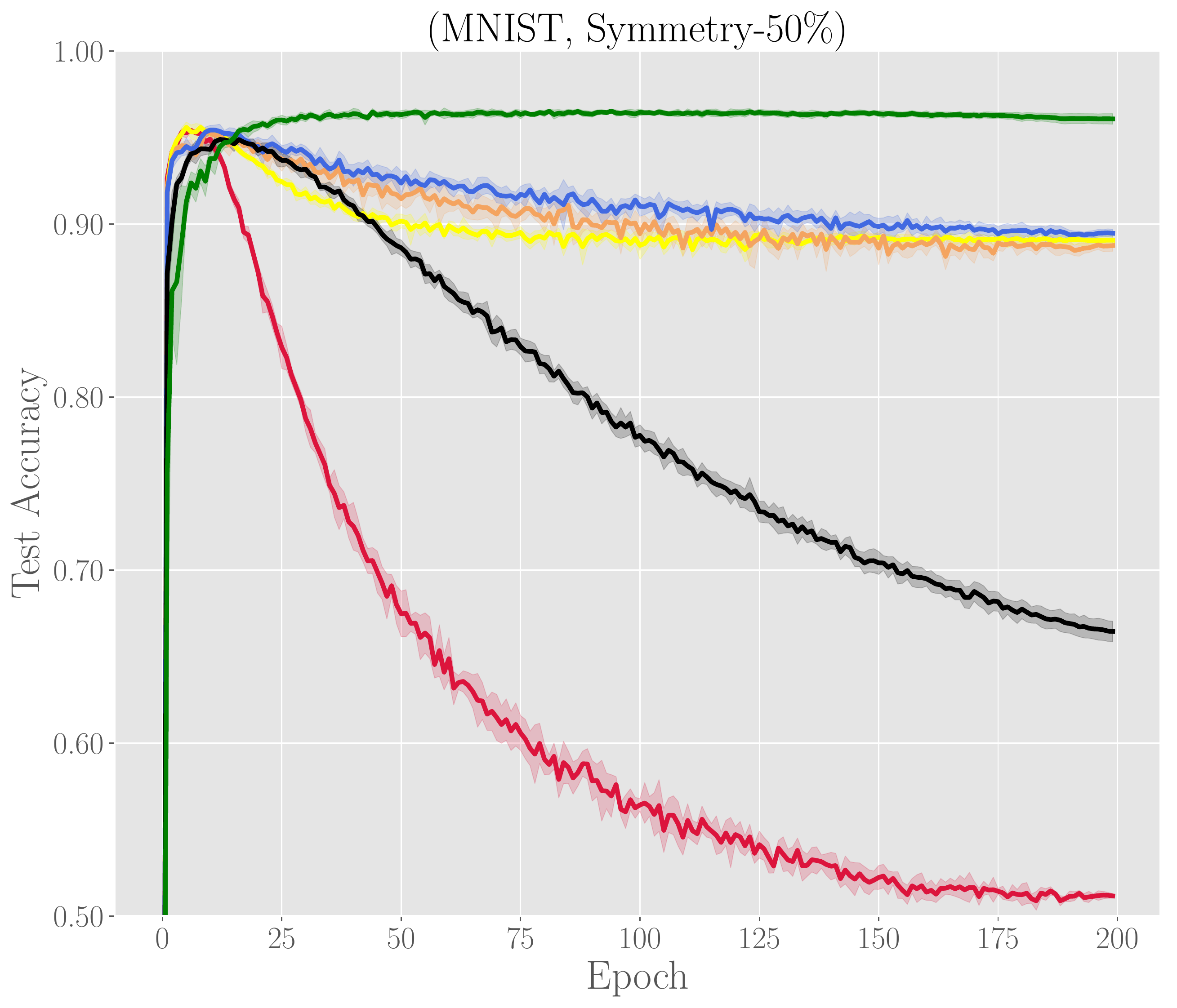}}
\subfigure[Symmetry-20\%.]
{\includegraphics[width=0.32\textwidth]{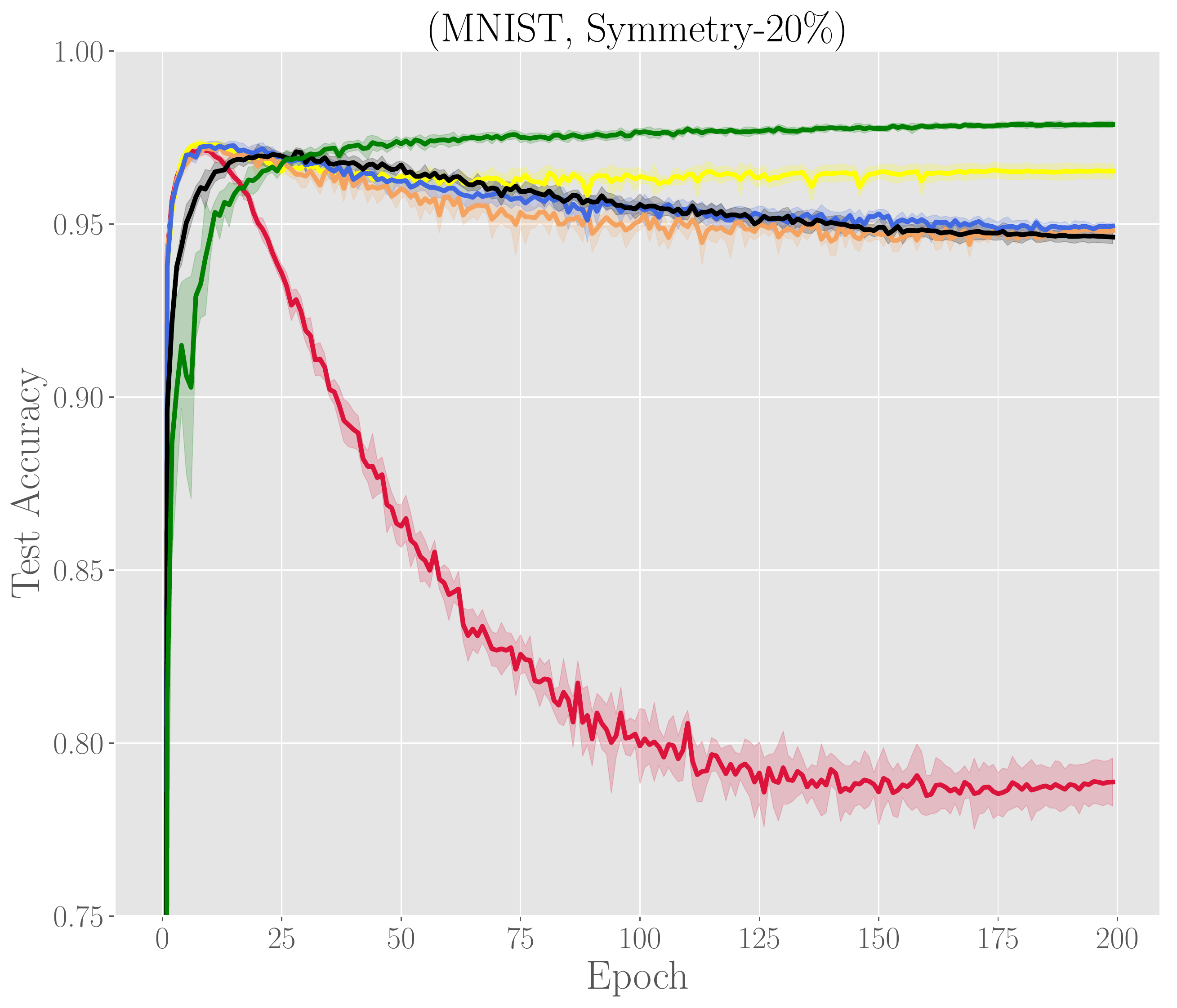}}
\caption{Test accuracy vs. number of epochs on \textit{MNIST} dataset.
    }
\label{fig:mnist}
%\vspace{-5px}
\end{figure*}

\paragraph{Results on \textit{MNIST}.} Figure~\ref{fig:mnist} shows
test accuracy vs. number of epochs on \textit{MNIST}.
In all three plots, we can clearly see the memorization effects of deep networks. For example, test accuracy of Standard first reaches a very high level since deep network will first fit clean labels. Over the increase of epochs, deep network will over-fit noisy labels gradually, which decreases its test accuracy accordingly. Thus, a robust training method should alleviate or even stop the decreasing trend in test accuracy.

In the easiest Symmetry-20\% case, all new approaches work better than Standard obviously, which demonstrates their robustness. Co-teaching+ and \mbox{F-correction} work significantly better than Co-teaching, MentorNet and Decoupling. However, F-correction cannot combat with the other two harder cases, i.e., Pair-45\% and Symmetry-50\%. Especially in the hardest Pair-45\% case, F-correction can learn nothing at all, which greatly restricts its practical usage in the wild. Besides, in two such cases, Co-teaching+ achieves higher accuracy than Co-teaching and MentorNet.

\begin{figure*}[ht]
\centering\stackunder{\includegraphics[width=0.9\textwidth]{legend.pdf}}{}
\subfigure[Pair-45\%.]
{\includegraphics[width=0.32\textwidth]{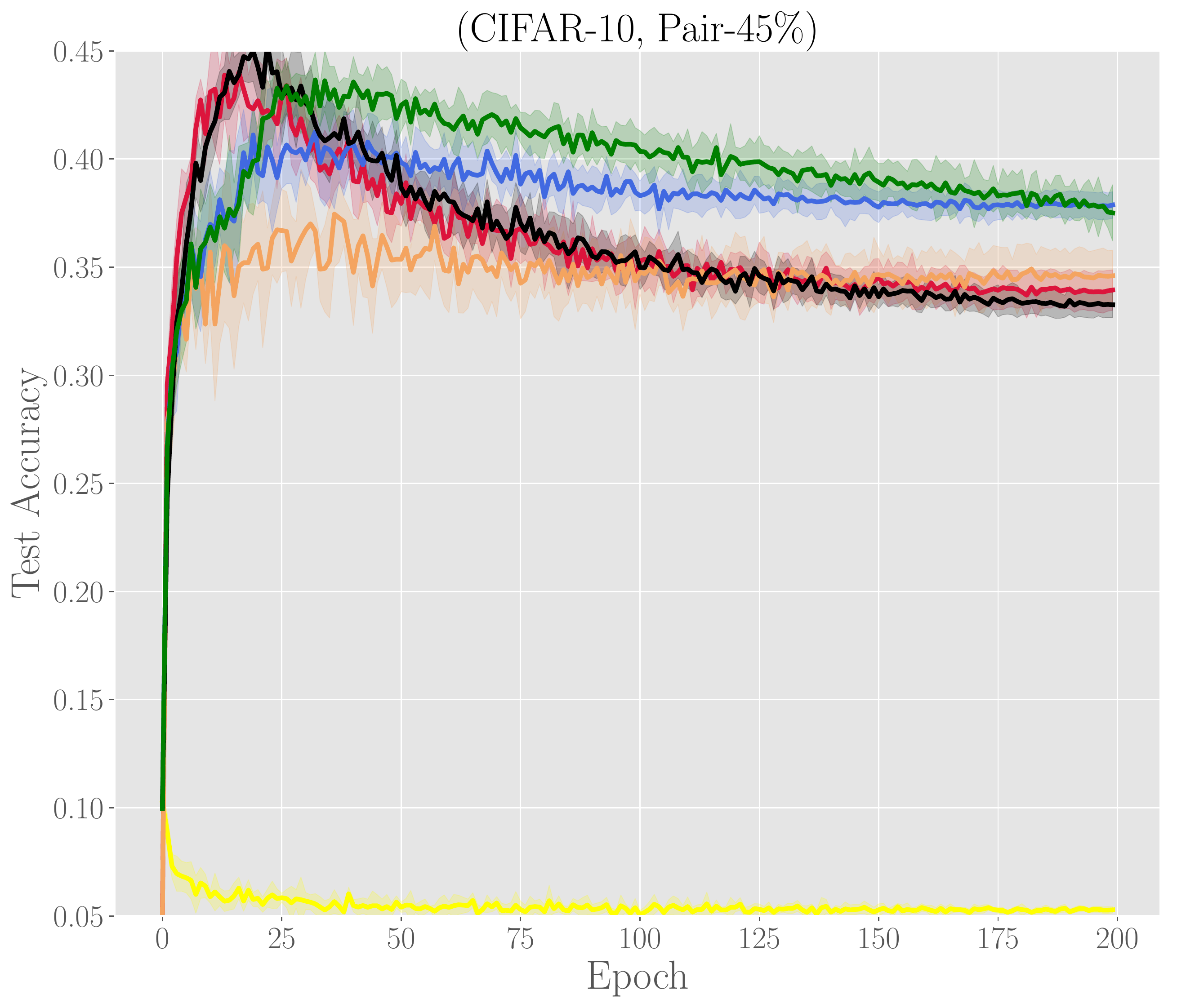}}
\subfigure[Symmetry-50\%.]
{\includegraphics[width=0.32\textwidth]{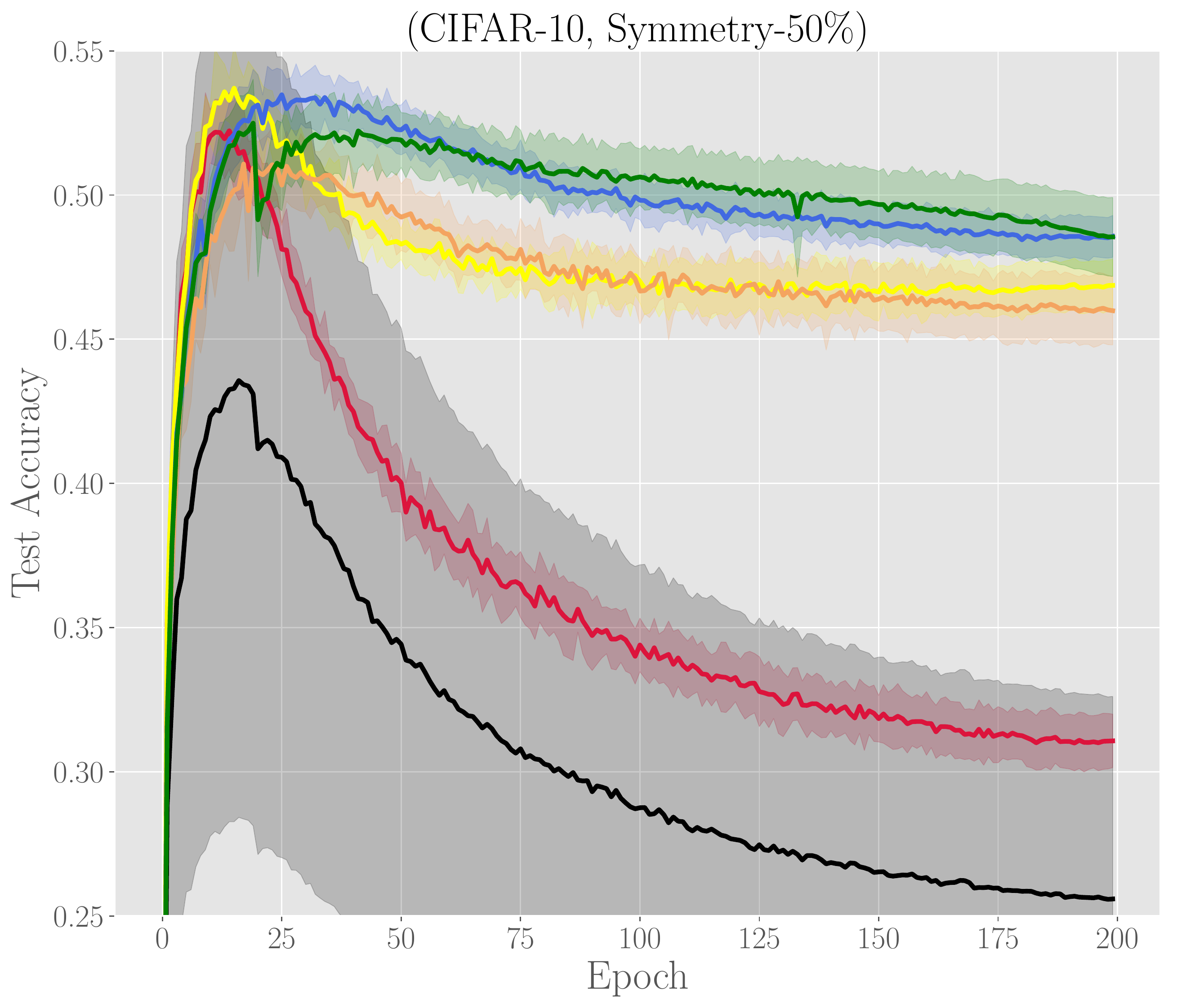}}
\subfigure[Symmetry-20\%.]
{\includegraphics[width=0.32\textwidth]{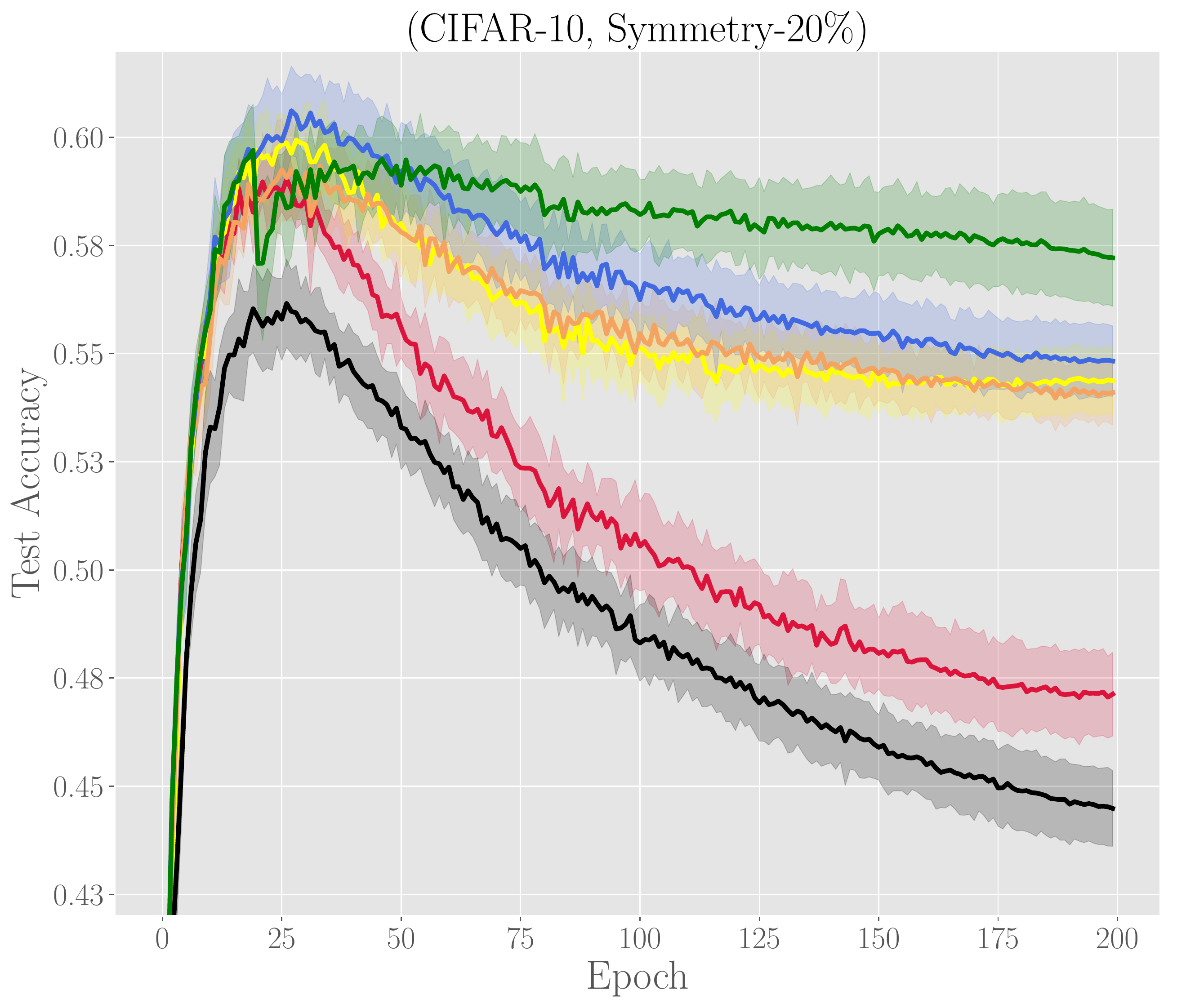}}
\caption{Test accuracy vs. number of epochs on \textit{CIFAR-10} dataset.
    }
\label{fig:cifar10}
%\vspace{-5px}
\end{figure*}

\textbf{Results on \textit{CIFAR-10}.}
Figure~\ref{fig:cifar10} shows
test accuracy vs. number of epochs on \mbox{\textit{CIFAR-10}}. Similarly, we can clearly see the memorization effects of deep networks, namely test accuracy of Standard first reaches a very high level then decreases gradually. In the easiest Symmetry-20\% case, Co-teaching+ works much better than all other baselines, where F-correction works similar to MentorNet but a bit worse than Co-teaching.

However, F-correction cannot combat with two harder cases easily, i.e., Pair-45\% and Symmetry-50\%. In the Symmetry-50\% case, F-correction works better than Standard and Decoupling, but worse than Co-teaching and Co-teaching+. In the hardest Pair-45\% case, F-correction almost learns nothing. In such two harder cases, our Co-teaching+ consistently achieves higher accuracy than Co-teaching and MentorNet.

%To explain such good performance, we plot label precision vs. number of epochs in the bottom of Figure~\ref{fig:cifar10}. Similarly, the higher label precision leads to the higher test accuracy during training. Compared to Standard and Decoupling, another three approaches can successfully pick clean instances out. However, our Co-teaching+ always achieve the highest label precision cross three cases, which demonstrates that our new approach is better at finding clean instances. Thus, the test accuracy of our Co-teaching+ is higher than others. In two easiest cases, the label precision of Co-teaching+ increases fast to a high plateau and then keep constant. This may explain why the test accuracy of Co-teaching performs better and better with the increase of epochs.

\begin{figure*}[ht]
\centering\stackunder{\includegraphics[width=0.9\textwidth]{legend.pdf}}{}
\subfigure[Pair-45\%.]
{\includegraphics[width=0.32\textwidth]{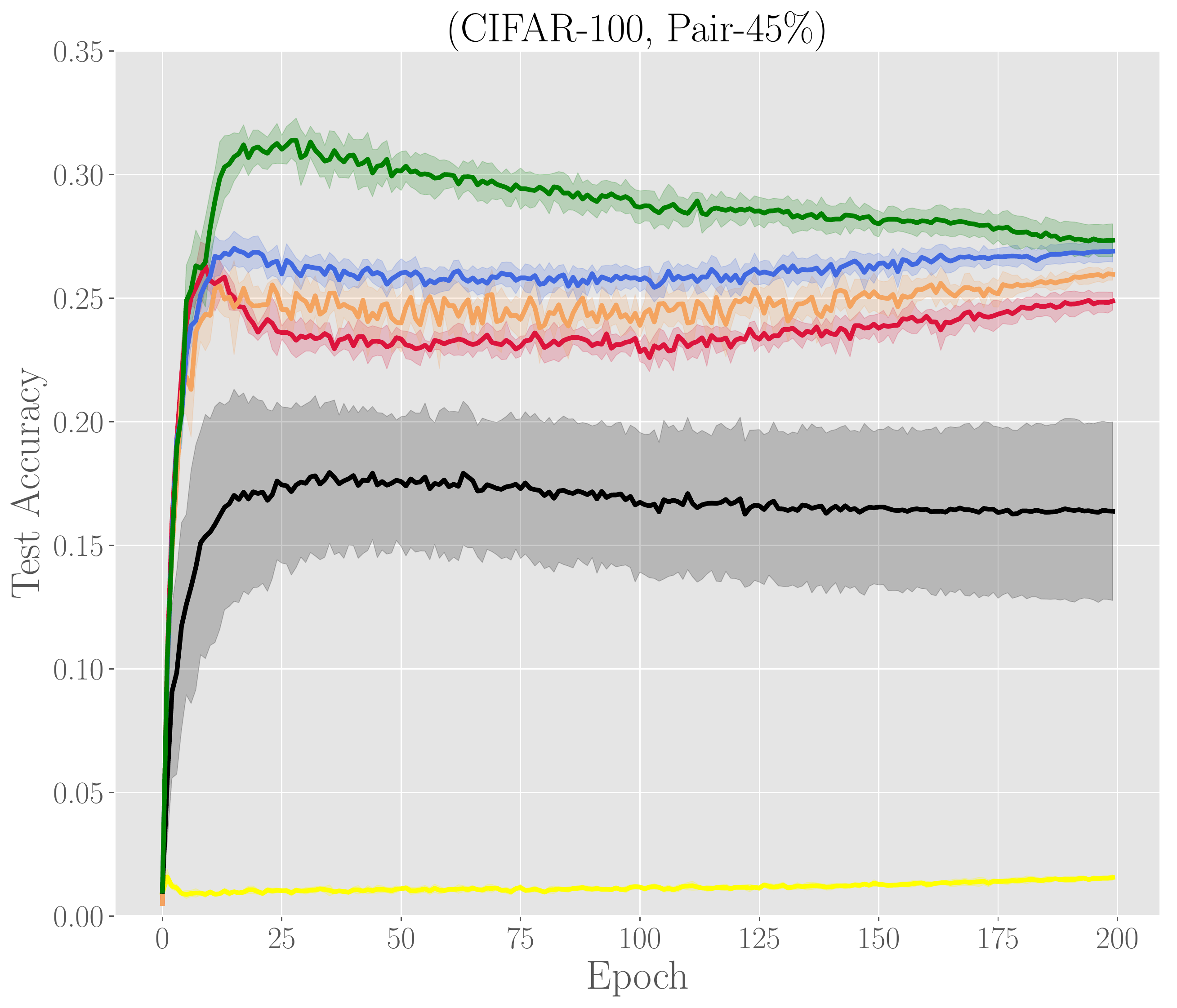}}
\subfigure[Symmetry-50\%.]
{\includegraphics[width=0.32\textwidth]{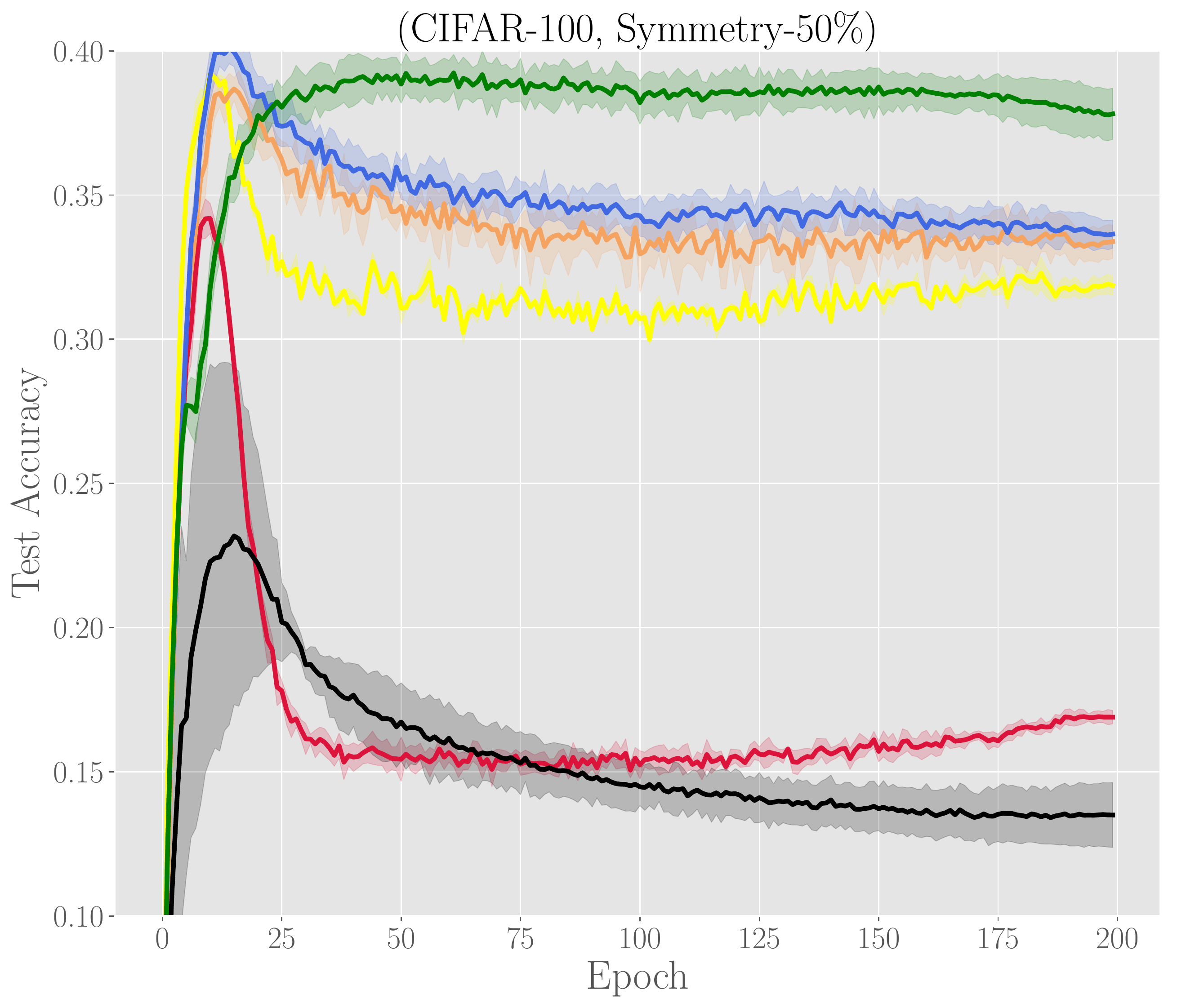}}
\subfigure[Symmetry-20\%.]
{\includegraphics[width=0.32\textwidth]{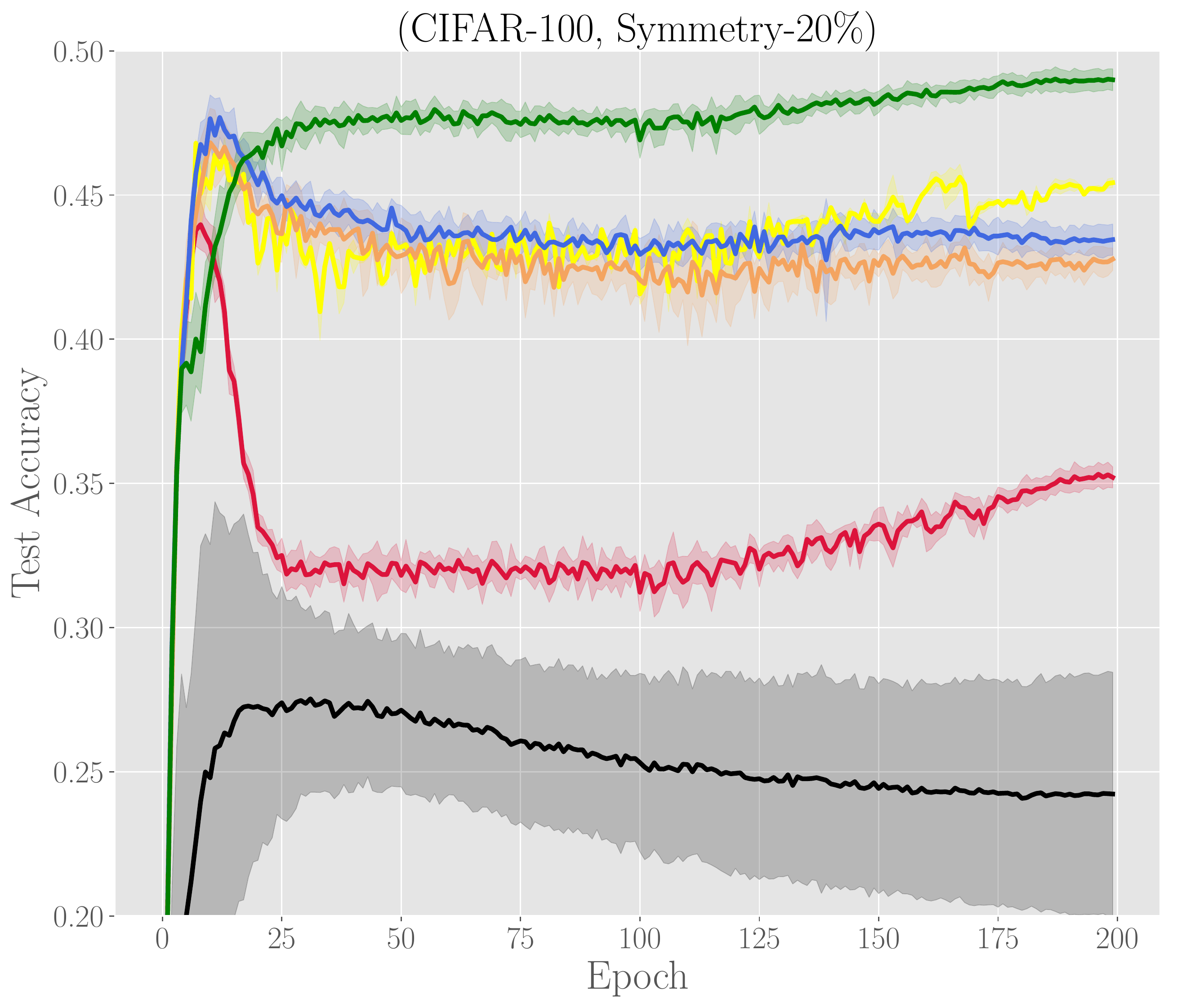}}
\caption{Test accuracy vs. number of epochs on \textit{CIFAR-100} dataset.
    }
\label{fig:cifar100}
\end{figure*}

\textbf{Results on \textit{CIFAR-100}.}
Figure~\ref{fig:cifar100} shows test accuracy vs. number of epochs on \mbox{\textit{CIFAR-100}}. Similarly, we can clearly see the memorization effects of deep networks, namely test accuracy of Standard first reaches a very high level then decreases gradually. In the easiest Symmetry-20\% case, Co-teaching+ and F-correction work significantly better than Co-teaching, MentorNet and Decoupling.

However, F-correction cannot combat with two harder cases easily, i.e., Pair-45\% and Symmetry-50\%. In the Symmetry-50\% case, F-correction works better than Standard and Decoupling, but worse than the other three approaches. In the hardest Pair-45\% case, F-correction almost learns nothing. In such two harder cases, our Co-teaching+ consistently achieves higher accuracy than Co-teaching and MentorNet. An interesting phenomenon is, in the easiest case, Co-teaching+ not only fully stop the decreasing trend in test accuracy, but also performs better and better with the increase of epochs.

\begin{figure*}[ht]
\centering\stackunder{\includegraphics[width=0.9\textwidth]{legend.pdf}}{}
\subfigure[Pair-45\%.]
{\includegraphics[width=0.32\textwidth]{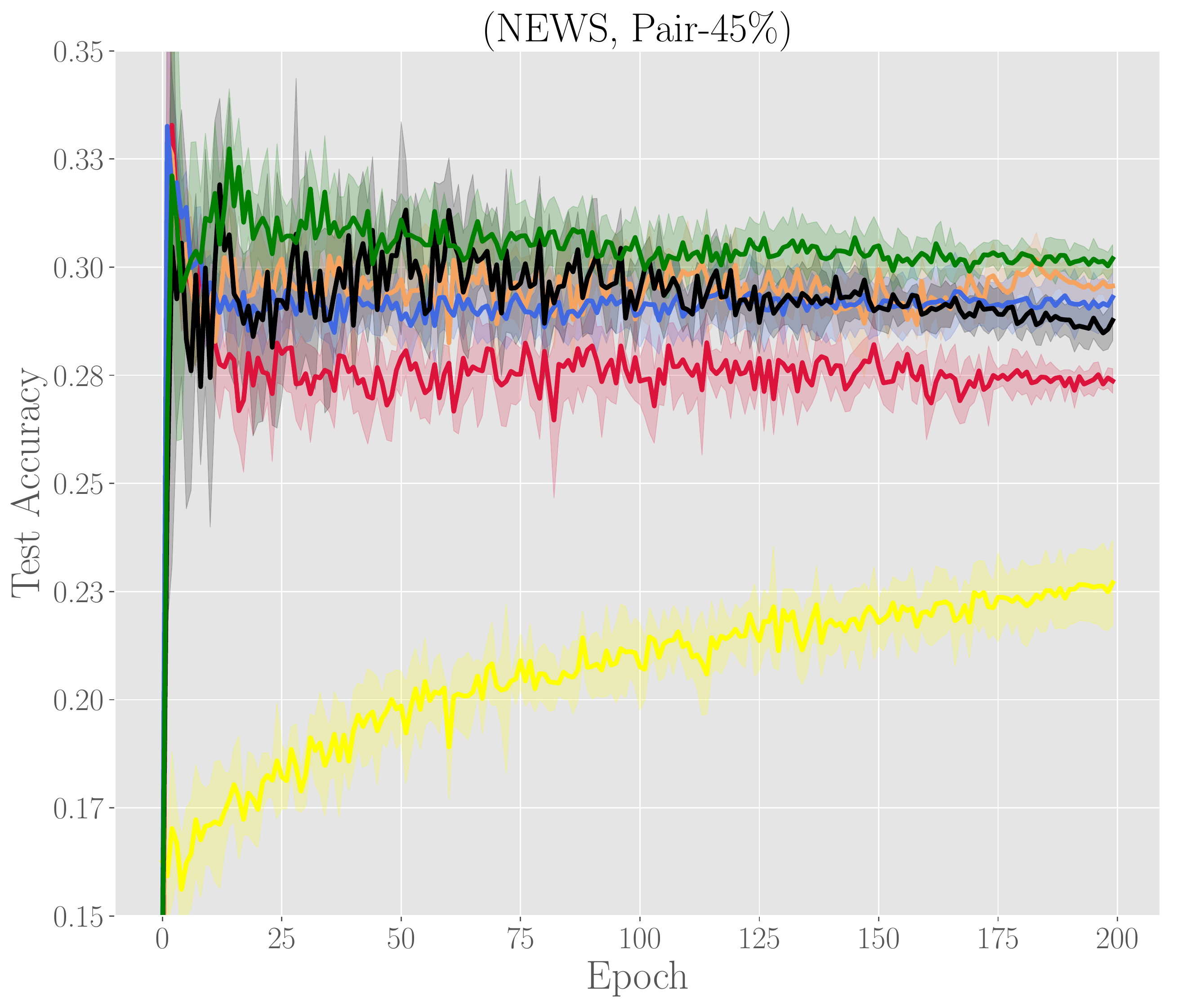}}
\subfigure[Symmetry-50\%.]
{\includegraphics[width=0.32\textwidth]{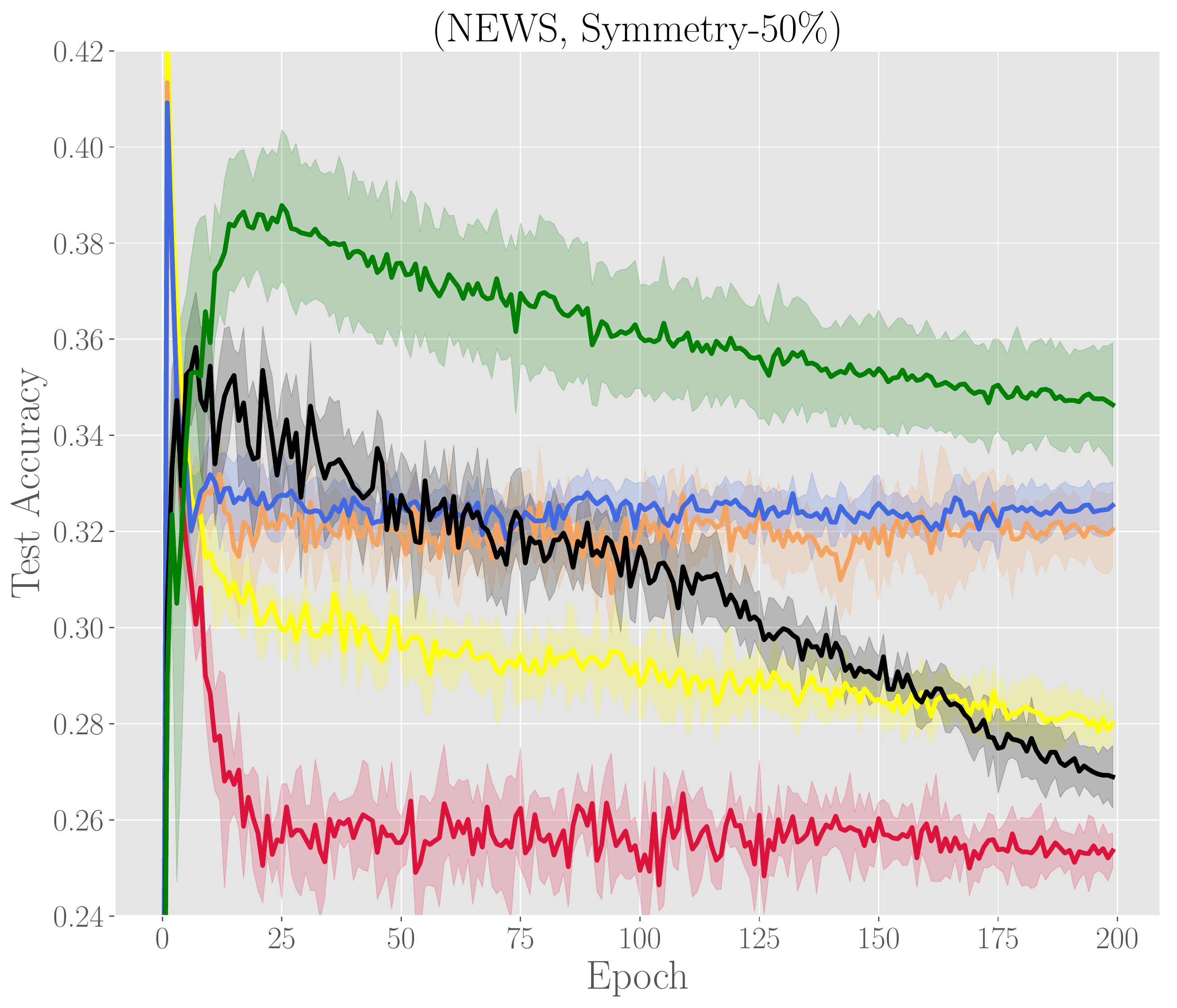}}
\subfigure[Symmetry-20\%.]
{\includegraphics[width=0.32\textwidth]{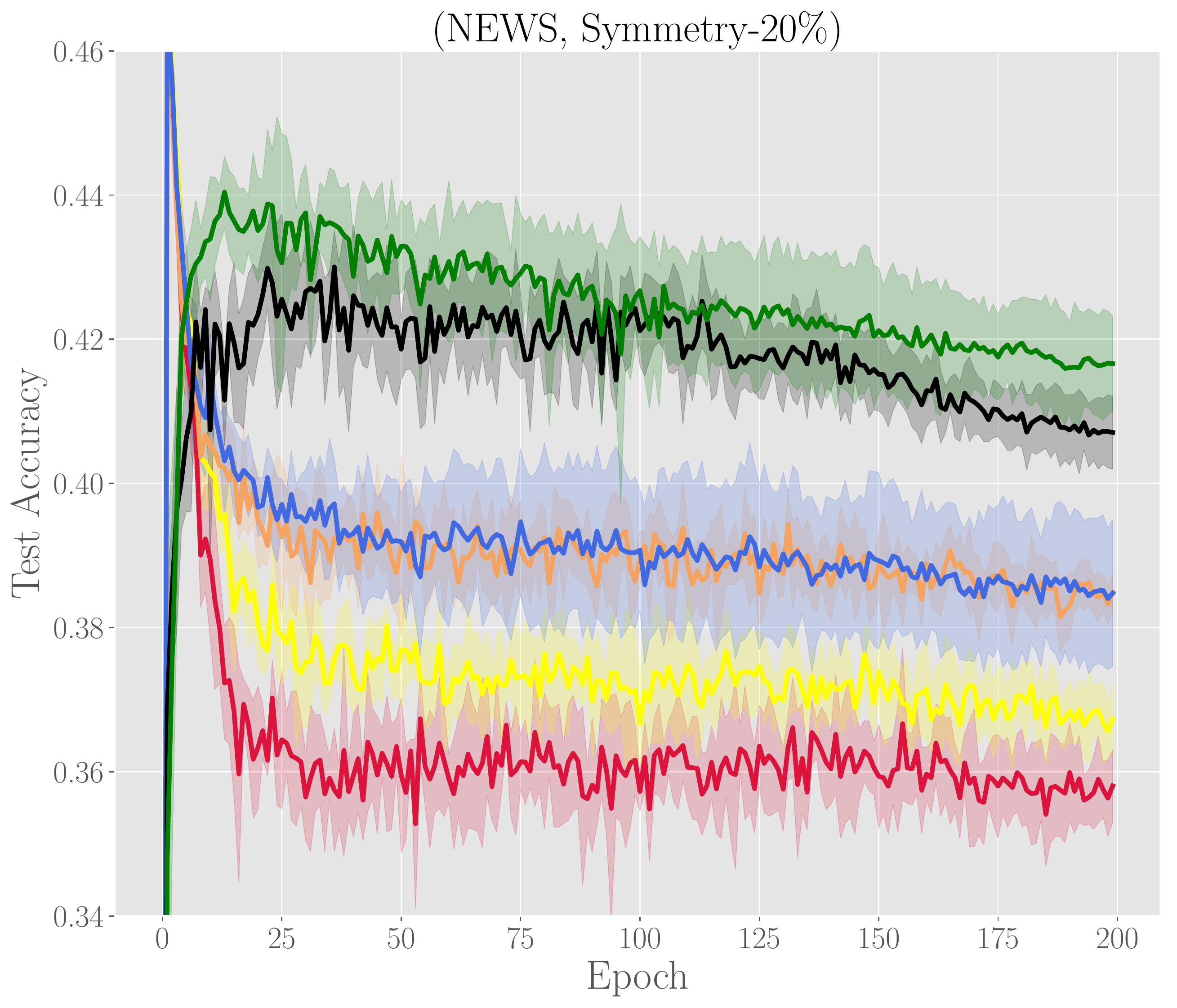}}
\caption{Test accuracy vs. number of epochs on \textit{NEWS} dataset.
    }
\label{fig:20news}
%\vspace{-5px}
\end{figure*}

\vspace{-5px}
\textbf{Results on \textit{NEWS}.}
To verify Co-teaching+ comprehensively, we conduct experiments not only on vision datasets, but also on text dataset \textit{NEWS}. Figure~\ref{fig:20news} shows
test accuracy vs. number of epochs on \textit{NEWS}.

Similar to results on vision datasets, we can still see the memorization effects of deep networks in all three plots, i.e., test accuracy of Standard first reaches a very high level and then gradually decreases. However, Co-teaching+ mitigates such memorization issue, and works much better than others across three cases. Note that F-correction cannot combat with all three cases, even in the easiest Symmetry-20\% case. This interesting phenomenon in F-correction does not occur in vision datasets.

%To explain such good performance, we plot label precision vs. number of epochs in the bottom of Figure~\ref{fig:20news}. Similarly, the higher label precision leads to the higher test accuracy during training. Compared to Standard and Decoupling, another three approaches can successfully pick clean instances out.
%However, in all three cases, Co-teaching+ always achieve the highest label precision, which demonstrates that our new approach is better at finding clean instances than Co-teaching and MentorNet. Thus, the test accuracy of our Co-teaching+ is higher than others.

%Noted that in the Pair-45\% case, the label precision of Co-teaching+ first increases to a high peak value, and then drops below that of Co-teaching and MentorNet. Nonetheless, the test accuracy of Co-teaching+ keep stable. The reason can be concluded in the following. Within around 180 epochs, the label precision of Co-teaching is always higher than Co-teaching and MentorNet, which trains a robust deep network. After 180 epochs, even when the label precision of Co-teaching+ is slightly lower than other two methods, the test accuracy of Co-teaching will not be affected since the robustness of deep network has already been established.

\textbf{Results on \textit{T-ImageNet}.}
To verify our approach on a complex scenario, Table~\ref{tab:imagenet_tiny} shows averaged/maximal test accuracy on \mbox{\textit{T-ImageNet}} over last 10 epochs. As we can see, for both Symmetry cases, Co-teaching+ is the best. For the Pair case, Co-teaching and Co-teaching+ outperform other four baselines.

\begin{table*}[!tp]\small
	\centering
	\caption{Averaged/maximal test accuracy (\%) of different approaches on \textit{T-ImageNet} over last 10 epochs. The best results are in bold.}
	\label{tab:imagenet_tiny}
	\scalebox{1}
	{
		\begin{tabular}{c | c | c | c | c | c | c }
			\hline
			Flipping-Rate(\%)     & Standard & Decoupling & F-correction & MentorNet & Co-teaching & Co-teaching+      \\ \hline
			Pair-45\%          & 26.14/26.32    & 26.10/26.61       & 0.63/0.67         & 26.22/26.61     & 27.41/\textbf{27.82}       & 26.54/26.87    \\ \hline
			Symmetry-50\%      & 19.58/19.77    & 22.61/22.81      & 32.84/33.12        & 35.47/35.76     & 37.09/37.60       & 41.19/\textbf{41.77}    \\ \hline
            Symmetry-20\%      & 35.56/35.80    & 36.28/36.97       & 44.37/44.50        & 45.49/45.74     & 45.60/46.36           & 47.73/\textbf{48.20}        \\ \hline
		\end{tabular}
	}
\end{table*}

\begin{table*}[!tp]\small
	\centering
	\caption{Averaged/maximal test accuracy (\%) of different approaches on \textit{Open-sets} over last 10 epochs. The best results are in bold.}
	\label{tab:openset}
	\scalebox{1}
	{
		\begin{tabular}{c | c | c | c | c | c }
			\hline
			Open-set noise                        & Standard  & MentorNet & Iterative \cite{wang2018iterative} & Co-teaching & Co-teaching+ \\ \hline
			\textit{CIFAR-10}+\textit{CIFAR-100}  & 62.92     & 79.27/79.33     & 79.28     & 79.43/79.58     & 79.28/\textbf{79.74}    \\ \hline
			\textit{CIFAR-10}+\textit{ImageNet-32} & 58.63     & 79.27/79.40     & 79.38     & 79.42/79.60     & 79.89/\textbf{80.52}    \\ \hline
            \textit{CIFAR-10}+\textit{SVHN}       & 56.44     & 79.72/79.81     & 77.73     & 80.12/80.33     & 80.62/\textbf{80.95}    \\ \hline
		\end{tabular}
	}
\end{table*}

\section{Experiments on Real-world Noisy Datasets}
\label{sect:experiments_openset}

\subsection{Experimental setup}\label{sec:setup}

\paragraph{Dataset.} To verify the efficacy of our approach in real-world scenario, we conduct experiments on open-set noisy datasets (abbreviated as \textit{Open-sets}) \cite{wang2018iterative}. Specifically, \textit{Open-sets} are built by replacing some training images in \textit{CIFAR-10} by outside images, while keeping the labels and the number of images per class unchanged. The ``mislabeled'' images come from different outside datasets, including \textit{CIFAR-100}, \textit{ImageNet-32} (32 $\times$ 32 ImageNet images) and \textit{SVHN}. Note that outside images whose labels exclude $10$ classes in \textit{CIFAR-10} are considered.

\paragraph{Network \& Optimizer \& Initialization.} We follow the experimental settings in \cite{wang2018iterative}. Specifically, we use a network architecture with $6$ convolutional layers and $1$ fully-connected layer, and its details can be found in the third column of Table~\ref{tab:netstuc}. Batch normalization (BN) is applied in each convolutional layer before the ReLU activation, and a max-pooling layer is implemented every two convolutional layers. All networks are trained by Stochastic Gradient Descent (SGD) with learning rate $0.01$, weight decay $10^{-4}$ and momentum $0.9$, and the learning rate is divided by $10$ after $40$ and $80$ epochs ($100$ in total).

Note that \textit{Open-sets} are real-world noisy datasets. To handle these complex scenarios, we should set the ratio of small-loss samples $\lambda(e)$ as follows.
\begin{equation}
    \lambda(e)=1-\min\{\frac{e}{E_{k}}\tau, (1+\frac{e-E_k}{E_{\max}-E_{k}})\tau\},
\end{equation}
where $E_k = 10$ and $E_{\max} = 200$.

\subsection{Comparison with the State-of-the-Arts}\label{sec:comp}

\textbf{Results on three \textit{Open-sets}.} Following \cite{wang2018iterative}, we report the classification accuracy on \textit{CIFAR-10} noisy datasets with 40\% open-set noise in Table \ref{tab:openset}. The Standard and Iterative results are borrowed from \cite{wang2018iterative}. For MentorNet, Co-teaching and Co-teaching+, we report the averaged/maximal test accuracy over the last 10 epochs. As can be seen, our approach outperforms other baselines on all three open-set noisy datasets. For \textit{CIFAR-100} noise and \textit{ImageNet-32} noise, both Co-teaching and Co-teaching+ are better than Iterative. For \textit{SVHN} noise, Co-teaching+ is significantly better than Iterative; while MentorNet and Co-teaching also work better than Iterative.

%\vspace{-5px}
\paragraph{Reflection of results.}
Different algorithm designs lead to different results. To sum up, self-paced MentorNet is concluded as training single deep network using the small-loss trick. Co-teaching moves further step, which is viewed as cross-training double deep networks using the small-loss trick. Based on Co-teaching, Co-teaching+ is regarded as cross-training double \textit{diverged} deep networks using the small-loss trick. Thus, keeping two deep networks diverged is one of the key ingredients to train robust deep networks. This point has been empirically verified by the result difference between Co-teaching and Co-teaching+.
%Safeguarded Dynamic Label Regression for Noisy Supervision
%However, all methods require either extra resources or more complex networks
%\footnote{+++ Adding some new papers here.}
%\vspace{-10px}
%\paragraph{Learning to teach methods.}
%Learning-to-teach is also a hot topic.
%Inspired by \cite{hinton2015distilling},
%these methods are made up by teacher and student networks.
%The duty of teacher network is to select more informative instances for better training of student networks.
%Recently,
%such idea is applied to learn a proper curriculum for the training data \cite{fan2017learning}
%and deal with multi-labels \cite{gong2016teaching}.
%However,
%these works do not consider noisy labels,
%and MentorNet \cite{jiang2018mentornet} introduced this idea into such area.

\section{Conclusion}
\label{sect:conclusion}
This paper presents a robust learning paradigm called Co-teaching+, which trains deep neural networks robustly under noisy supervision. Our key idea is to maintain two networks simultaneously that find the prediction disagreement data. Among such disagreement data, our method cross-trains on data screened by the ``small loss'' criteria. We conduct experiments to demonstrate that, our proposed Co-teaching+ can train deep models robustly with the extremely noisy supervision beyond Co-teaching and MentorNet. More importantly, we summarize three key points towards training robust deep networks with noisy labels: (1) using small-loss trick based on memorization effects of deep networks; (2) cross-updating parameters of two networks; and (3) keeping two deep networks diverged during the whole training epochs.
%In future,
%we can extend our work in the following aspects. First, we can improve the simple disagreement rule using curriculum learning, which may speed up the convergence rate.
%Second, we can adapt Co-teaching+ paradigm to train deep models under other weak supervisions, e.g., unlabeled and unlabeled data \cite{lu2018minimal}.
In future, we will investigate the theory of Co-teaching+ from the view of disagreement-based algorithms
\cite{wang2017theoretical}.
%Since there is no analysis for generalization performance on deep learning with noisy labels, we leave the generalization analysis as a future work.

\section*{Acknowledgments}
MS was supported by JST CREST JPMJCR18A2. IWT was supported by ARC FT130100746, DP180100106 and LP150100671. XRY was supported by China Scholarship Council No. 201806450045. We gratefully acknowledge the support of NVIDIA Corporation with the donation of Titan Xp GPU used for this research.
\bibliography{coteachingplus.bbl}
\bibliographystyle{icml2019}

\appendix
\cleardoublepage
\onecolumn

\section{Related literature}
\label{sect:relatedworks}
\paragraph{Statistical learning methods.}
Statistical learning contributed a lot to the problem of learning with noisy labels, especially in theoretical aspects. Statistical learning approaches can be categorized into three strands: surrogate loss, noise rate estimation and probabilistic modeling. For example, in the surrogate losses category,
\citet{natarajan2013learning} proposed an unbiased estimator to provide the noise corrected loss approach.
\citet{masnadi2009design} presented a robust non-convex loss, which is the special case in a family of robust losses.
In the noise rate estimation category, both \citet{menon2015learning} and \citet{liu2016classification} proposed a class-probability estimator using order statistics on the range of scores. \citet{sanderson2014class} presented the same estimator using the slope of the ROC curve. In the probabilistic modeling category, \citet{raykar2010learning} proposed a two-coin model to handle noisy labels from multiple annotators. \citet{yan2014learning} extended this two-coin model by setting the dynamic flipping probability associated with instances.
\vspace{-10px}
\paragraph{Deep learning approaches.}

Deep learning approaches are prevalent to handle noisy labels \cite{zhang2018generalized}.
\citet{li2017learning} proposed a unified framework to distill the knowledge from clean labels and knowledge graph, which can be exploited to learn a better model from noisy labels.
\citet{veit2017learning} trained a label cleaning network by a small set of clean labels, and used this network to reduce the noise in large-scale noisy labels.
\citet{rodrigues2017deep} added a crowd layer after the output layer for noisy labels from multiple annotators.
\citet{tanaka2018joint} presented a joint optimization framework to learn parameters and estimate true labels simultaneously.
\citet{ren2018learning} leveraged an additional validation set to adaptively assign weights to training examples. Similarly, based on a small set of trusted data with clean labels,
\citet{hendrycks2018using} proposed a loss correction approach to mitigate the effects of label noise on deep neural network classifiers.
\citet{ma2018dimensionality} developed a new dimensionality-driven learning
strategy, which monitors the dimensionality of deep representation subspaces during training and adapts the loss function accordingly.
\citet{wang2018iterative} proposed an iterative learning framework for training
CNNs on datasets with open-set noisy labels.
\citet{han2018masking} proposed a human-assisted approach that conveys human cognition of invalid class transitions, and derived a structure-aware deep probabilistic model incorporating a speculated structure prior.
\citet{lee2019robust} proposed a novel inference method to obtain a robust decision boundary under any softmax neural classifier pre-trained on noisy datasets. Their idea is to induce a generative classifier on top of hidden
feature spaces of the discriminative deep model.

\section{Training details}
For \textit{MNIST} and \textit{NEWS}, we train Co-teaching+ by default at the beginning of training. For other datasets, we use a warm-up strategy to achieve a higher test accuracy. Specifically, for \textit{CIFAR-10}, we warm-up Co-teaching+ with training Co-teaching for the first $20$ epochs (i.e., only conducting cross-update for the first $20$ epochs). For \textit{CIFAR-100}, we warm-up Co-teaching+ with training Co-teaching for the first $5$ epochs. For \textit{T-ImageNet}, we start disagreement-update in the middle of training, i.e., we warm-up Co-teaching+ with training Co-teaching for the first $100$ epochs. For \textit{Open-sets}, we warm-up Co-teaching+ with training two networks in parallel for the first $55$ epochs, where both networks leverage the small-loss trick. Inevitably, there is few chance that we cannot find enough small-loss instances for cross-update. In that case, we only conduct disagreement-update in a mini-batch data during training.
\end{document}